\pdfoutput=1
\documentclass[11pt]{article}

\usepackage{authblk}
\usepackage{naacl2021}

\usepackage{times}
\usepackage{latexsym}
\usepackage{pifont}% http://ctan.org/pkg/pifont
\newcommand{\cmark}{\ding{51}}%
\newcommand{\xmark}{\ding{55}}%

\usepackage[T1]{fontenc}
\usepackage[utf8]{inputenc}

\usepackage{microtype}

\usepackage{times}
\usepackage{latexsym}
\usepackage{graphicx}
\usepackage{amssymb}
\usepackage{hyperref}
\usepackage[capitalise]{cleveref}
\usepackage{tabularx}
\usepackage{tablefootnote}
\usepackage{subcaption}
\usepackage{makecell}
\usepackage{booktabs}
\usepackage{multirow}
\usepackage{ulem}
\usepackage{soul}

\newcommand{\nop}[1]{}
\newcommand{\hs}[1]{#1}

\newcommand{\add}[1]{#1}

\newcommand{\ours}[0]{\textsc{StruG}}
\newcommand{\newdata}[0]{Spider-Realistic\ }

\title{Structure-Grounded Pretraining for Text-to-SQL}

\author[1]{Xiang Deng \thanks{Work done during an internship at Microsoft Research.}}
\author[2]{Ahmed Hassan Awadallah}
\author[2]{Christopher Meek}
\author[2]{\\Oleksandr Polozov}
\author[1]{Huan Sun}
\author[2]{Matthew Richardson}

\affil[1]{%
The Ohio State University}
\affil[ ]{\texttt {\{deng.595,sun.397\}@osu.edu}}
\affil[2]{%
Microsoft Research, Redmond}
\affil[ ]{\texttt {\{hassanam,meek,polozov,mattri\}@microsoft.com}}

\begin{document}
\maketitle
\begin{abstract}
% \textcolor{red}{Candidate titles:
% \begin{itemize}
%     \item Pretraining beyond Masking: Weakly Supervised Multi-Task Objectives for Pretraining Text-to-SQL
%     \item Weakly Supervised Multi-Task Pretraining for Text-to-SQL
% \end{itemize}}
% \matt{is A4 size correct?}\xd{It comes with the template so should be correct}
Learning to capture text-table alignment is essential for\nop{table}  tasks like text-to-SQL. A model needs to correctly recognize natural language references to columns and values and to ground them in the given database schema. In this paper, we present a novel weakly supervised \textbf{Stru}cture-\textbf{G}rounded pretraining framework (\ours) for text-to-SQL that can effectively learn to capture text-table alignment based on a parallel text-table corpus\nop{extract text-table alignment knowledge from a parallel text-table corpus}. We identify a set of novel {pretraining}\nop{prediction} tasks: column grounding, value grounding and column-value mapping, and leverage them to pretrain a text-table encoder\nop{without requiring complex SQL annotation}. Additionally, to evaluate different methods\nop{the model} under more realistic text-table alignment settings, we create a new evaluation set \newdata based on  Spider {dev set} with explicit mentions of column names removed, and adopt eight existing\nop{single-database} text-to-SQL datasets {for cross-database evaluation}. \ours\ brings significant improvement over BERT\textsubscript{LARGE} in all settings. Compared with existing pretraining methods such as GRAPPA, \ours\ achieves similar performance on Spider, and outperforms all baselines on more realistic sets. The Spider-Realistic dataset is available at \url{https://doi.org/10.5281/zenodo.5205322}.\nop{All the code and data used in this work is public available at \url{https://aka.ms/strug}.}\nop{\ours \ significantly outperforms \hs{existing pretraining methods such as} BERT\textsubscript{LARGE} on Spider and {more}\nop{the} realistic evaluation sets, as well as brings consistent improvement on the large-scale WikiSQL benchmark. \nop{Huan: mention comparison with Grappa and TaBert? The last sentence may need to be re-summarized.}}
% \todo{REWRITE LATER BASED ON INTRO} Language models (LMs) pretrained on massive, heterogeneous text corpus constitute the basis for modern NLP models. Owing to the superior performance of these general purposed models, how to efficiently adopt them to the target task has attracted increasing attention recently. In this paper, we present a weakly supervised multi-task fine-tuning strategy for adopting LMs to the text-to-SQL task. XXX leverages large amount of weakly supervised table-sentence pair data to fine-tune the pretrained language model (PLM) using three subtasks of text-to-SQL: value prediction, where column selection and value column mapping. In the experiments, XXX brings consistent improvement on complex and cross-domain text-to-SQL task. We also demonstrate using WikiSQL that XXX allows training a text-to-SQL model with substantially fewer annotations than directly adopting the PLM representations.
\end{abstract}
\section{Introduction}
Semantic parsing is the task of mapping a natural language (NL) utterance to a machine-understandable representation such as lambda calculus, abstract meaning representation, or a structured query language (e.g., SQL). In this paper, we focus on the task of translating NL questions to executable SQL queries (text-to-SQL). This is a fundamental task for building natural language interfaces for databases, which can enable non-expert users to effortlessly query databases \cite{Androutsopoulos1995NaturalLI,Li2014NaLIRAI}.

% Remarkable progress has been made in text-to-SQL within the past few years. With sufficient in-domain training data, existing models already achieve nearly 90\% accuracy on single-domain benchmarks like ATIS and GeoQuery. However, annotating NL questions with SQL queries is expensive and it is cost-prohibitive to collect training examples for all possible databases. A model that can generalize across domains and databases is desired. In light of this, \citeauthor{yu2018spider} present Spider, a cross-domain text-to-SQL benchmark that trains and evaluates the system using different databases, and has attracted much attention recently.

\begin{figure}[t]
\centering
\includegraphics[width=\linewidth]{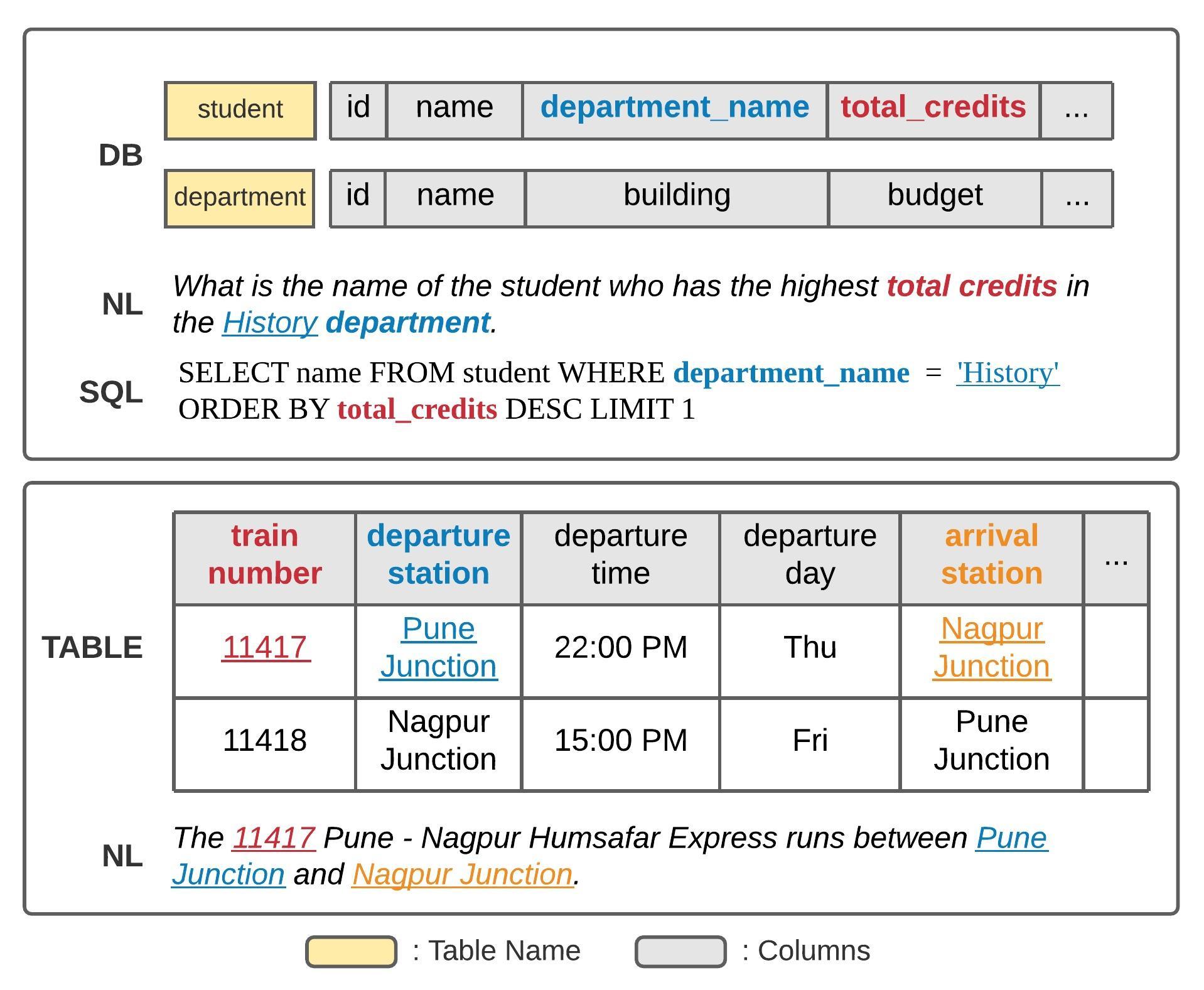}
\vspace{-15pt}
\caption{Illustration of text-to-SQL text-table alignment (top half) and parallel text-table corpus (bottom half). In both examples, the associations between tokens in the NL utterance and columns in the table are indicated. In this paper, we aim to leverage the text-table alignment knowledge in the parallel text-table corpus to help text-to-SQL. \nop{We can see that an essential challenge in text-to-SQL is to \textit{correctly identify natural language references to columns and values and to ground them in the given database schema}. As we can see from the bottom half of the figure, such column/value mention and alignment also naturally exist in a parallel text-table corpus. In this work, we aim to learn text-table association knowledge from the parallel corpus via pretraining and use it to help the downstream text-to-SQL task.}}
\vspace{-15pt}
\label{fig:illustration}
\end{figure}

% \todo{Remove. Instead of talking about the two components but only solve one, refer to Figure 1 and talk about our intuition of using the association knowledge to help text2sql} Most existing systems divide text-to-SQL into two problems: generating the high level query structure, and selecting columns and values to fill into the sketch. In cross-domain setting, the former requires the model to adapt to unique query structures and dataset conversions which is hard to achieve without a few in-domain samples, while the latter examines the model's ability to handle language and database variation across domains \cite{suhr2020exploring}. In this paper, we aim to enhance the model's capability of jointly understanding the NL utterance and database schema via pretraining using parallel corpus containing both NL sentences and tabular data, and consequently improve the second component.

% \citeauthor{suhr2020exploring} identifies three generalization challenges for cross-domain semantic parsing: language variation across domains, novel database and query structures and dataset conventions. In this paper we focus on the first challenge, as it evaluates the model's ability to understand both the NL utterance and DB schema across domains, and capture the correlation between them.

One of the key challenges in text-to-SQL is text-table alignment, that is, \textit{to correctly recognize natural language references to columns and values and to ground them in the given database schema}. Consider the example in the top half of \cref{fig:illustration}. A model needs to first identify the column mentions \texttt{\small total credits}, \texttt{\small department}, and value mention \texttt{\small History}, and then ground them to the given schema. This is challenging for three reasons. First, the model needs to jointly understand the NL utterance and the database schema, as the user may refer to a column using various expressions which usually differ from the original column name. Second, the model needs to be able to generalize to new database schemas and referential language that is not seen in training. Finally, in the case that \nop{using}{accessing} cell values is not possible, the model still needs to identify potential value mentions and link them to the correct columns without exhaustively searching and matching over the database.

{On the other hand,} text-table alignment\nop{also} naturally exists in parallel text-table corpora, e.g., web tables with context \cite{lehmberg2016large}, table-to-text generation datasets \cite{parikh2020totto,chen2020logical}, table-based question answering datasets \cite{pasupat2015compositional,chen2020hybridqa}. Such datasets can be collected from web pages, documents, etc., and requires much less human effort to create compared with text-to-SQL datasets. The bottom half of \cref{fig:illustration} gives an example of such an alignment dataset. There are three value mentions \texttt{\small 11417}, \texttt{\small Pune Junction} and \texttt{\small Nagpur Jnction}, which can be grounded to the \texttt{\small train number}, \texttt{\small departure station} and \texttt{\small arrival station} columns respectively. Such alignment information can be easily obtained by leveraging the table contents or using some human annotation. In this work, we aim to \nop{take advantage of} incorporate the text-table alignment knowledge contained in a parallel corpus via pretraining and use it to help the downstream text-to-SQL task.

% Pretrained language models like BERT have shown promising results on improving the generalization ability of natural language processing (NLP) models on many tasks. However, as these general language models are usually trained with only unstructured text. They do not have knowledge about encoding structured data like database schema, and more importantly how to align them with the NL utterances. To solve this, TaBERT \cite{yin2020tabert} and TAPAS \cite{herzig2020tapas} try to further pretrain the language model leveraging a large amount of web tables and their textual context. However, they still use the masked language model (MLM) objective which is both inefficient to train and weak at capturing the correlation between the text and tabular data. Closest to our work, Grappa \cite{yu2020grappa} incorporates a SQL semantic prediction (SSP) objective and pretrains it using synthetic question-SQL pairs constructed from via a synchronous context free grammar induced from existing text-to-SQL datasets. Though effective, Grappa mainly learns compositional inductive bias from the synthetic data which requires SQL annotations and cannot fully utilize the unsupervised text-table parallel data.

\begin{figure}[t]
    \centering
    \includegraphics[width=\linewidth]{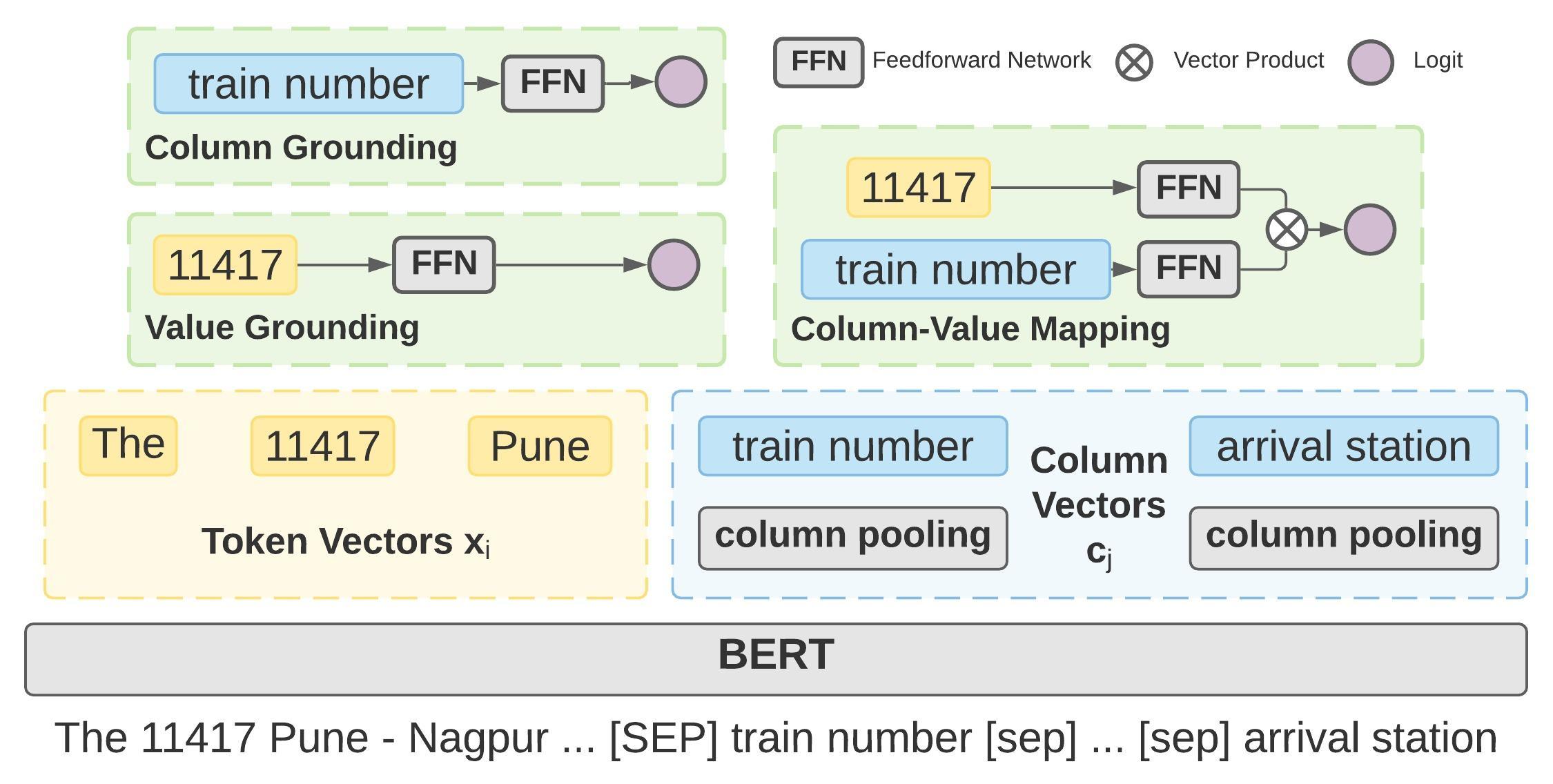}
    \vspace{-20pt}
    \caption{Overview of our model architecture and three pretraining objectives. \nop{seems a bit blurry.} }
    \vspace{-15pt}
    \label{fig:objective}
\end{figure}

\begin{figure*}[t]
\centering
\includegraphics[width=0.9\linewidth]{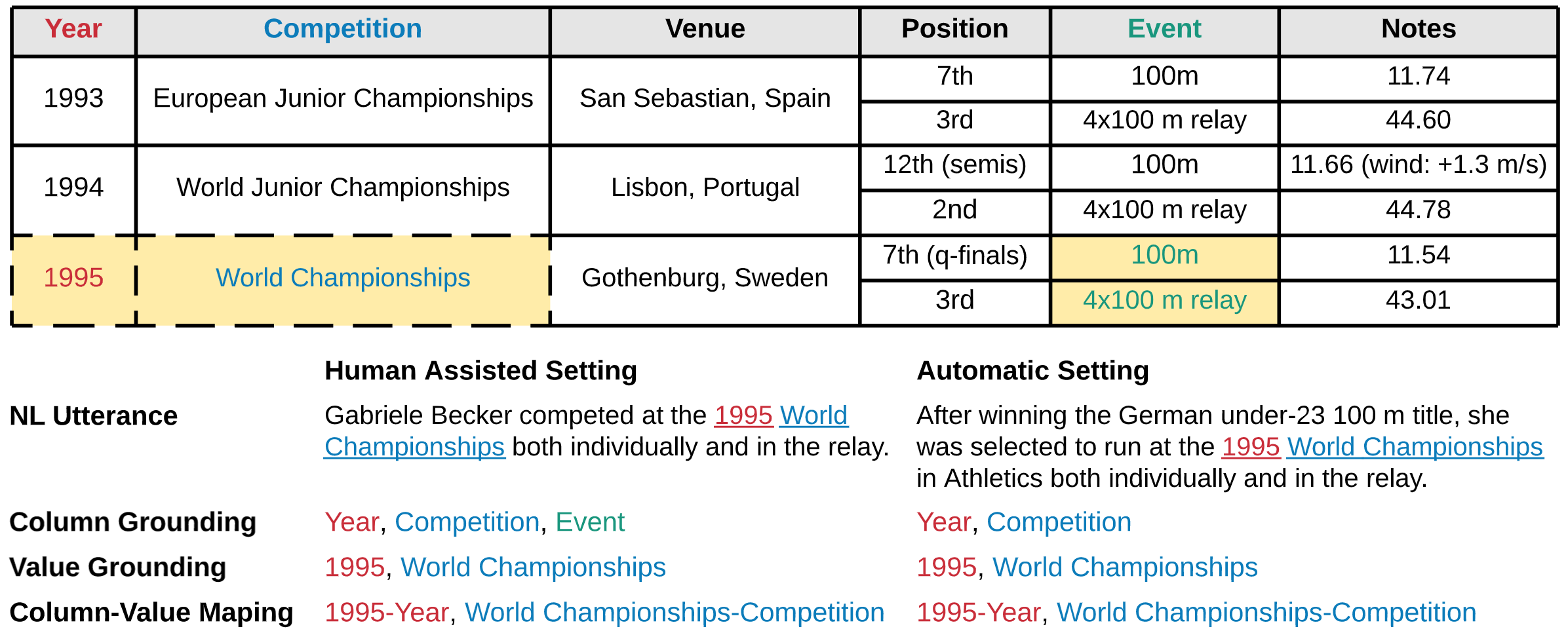}
\vspace{-5pt}
\caption{Illustration of the parallel corpus ToTTo \cite{parikh2020totto} and our two weakly supervised pretraining settings. Cell highlighted with yellow are the cell
annotations provided by ToTTo, and cell highlighted with dashed lines are cell annotations obtained via string matching in automatic setting.\nop{add row header NL}}
\vspace{-15pt}
\label{fig:pretraining}
\end{figure*}

We present a novel weakly supervised structure-grounded pretraining framework ({\ours}) for text-to-SQL. We design a set of prediction tasks and optimize them leveraging a parallel corpus containing both NL sentences and tabular data to encourage the encoded representation to capture information required to support tasks that require table grounding\nop{Huan: Should we refer to Figure 2 here?}. More specifically, we identify three critical tasks for aligning text with table: column grounding, value grounding and column-value mapping (examples shown in \cref{fig:objective}). We re-purpose an existing large-scale table-to-text generation dataset ToTTo \cite{parikh2020totto} for pretraining and gain labels for the three tasks via weak supervision\nop{setting (1) cannot be called weak supervision for these 3 tasks, right?}\nop{Yes, in both setting we are using noisy labels which is weak supervision}.\nop{ToTTo contains 120k NL descriptions and corresponding web tables automatically collected from Wikipedia using heuristics. It also provides descriptions revised by human annotators and cell level annotation of text-table association as illustrated in Figure \ref{fig:pretraining}.} We experiment under two settings, {with or without human assistance}\nop{with different levels of supervision}: (1) \textit{human assisted setting}, using ToTTo's revised descriptions and cell annotations; (2) \textit{automatic setting}, using the raw sentences and inferring the cell correspondences {via string matching with the table contents.}\nop{Huan: how about column mentions?} \nop{leveraging the table contents via string matching.} \nop{Although the human assisted setting still requires cell level supervision, such annotation is much easier to get compared with SQL queries while still offers high quality text-table alignment information. In the automatic setting, the simple heuristics may introduce more noisy labels but it can be easily extended to larger scale or domain specific parallel text-table corpus with no annotation needed.}

% \matt{The introduction need to be shorter. It's hard for the reader to follow because you are trying to make it concise but cover lots of specifics at the same time. It's okay to skip specifics save some of this for later. }
% \matt{I think it would be good to have the main challenges (that you have a pretraining objective for) appear early in the paper, to be more specific about what problems this method aims to solve. This should not be relative to Spider, but just stated generally like "One of the problems that text-to-SQL systems must solve is matching question terms to names of columns in the database" or such.}

% As pointed out by \citet{suhr2020exploring}, existing text-to-SQL benchmarks like Spider \cite{yu2018spider} ease the text-table grounding\hs{alignment, to be consistent?} challenge by using NL utterances that closely match their paired SQL query. \citet{suhr2020exploring} propose a new setting that includes 8 other single-domain text-to-SQL datasets for evaluation. However, besides text-table alignment, some of the datasets also contain challenges like novel query structure and dataset conventions. To give a more fair assessment of how well the joint representation captures the association between the text and database schema, aside from using some of the datasets from \citet{suhr2020exploring}, we create a new evaluation set from the original Spider development set by removing the explicit mention of column names in the utterance. This way we can overcome the previous mentioned limit of Spider without introducing new challenges.

As pointed out by \citet{suhr2020exploring}, existing text-to-SQL benchmarks like Spider \cite{yu2018spider} render the text-table alignment challenge easier than expected by explicitly mentioning exact column names in the NL utterances. Contrast this to more realistic settings where users may refer to the columns using a variety of expressions. \citet{suhr2020exploring} propose a new cross-database setting that uses Spider for training and includes eight other single-domain text-to-SQL datasets for evaluation. In addition to \nop{partly} adopting their setting, we create a new evaluation set called \newdata from the original Spider dev set, by removing explicit mentions of column names from an utterance. 
% \nop{However, besides text-table alignment, some of the datasets also contain challenges like novel query structure and dataset conventions. To give a more fair assessment of how well the joint representation captures the association between the text and database schema, aside from using some of the datasets from \citet{suhr2020exploring}, we create a new evaluation set from the original Spider development set by removing the explicit mention of column names in the utterance. This way we can overcome the previous mentioned limit of Spider without introducing new challenges.}

We pretrain \ours \nop{is our model structure the same as BERT\textsubscript{LARGE}?}\nop{yes} using 120k text-table pairs from ToTTo. Experiments show that our structure-grounded pretraining objectives are very efficient and usually converge with around 5 epochs \add{in less than 4 hours}. This dramatically reduces the pretraining cost compared to previous pretraining methods \cite{herzig2020tapas,yin2020tabert}. We adopt the same model architecture as BERT \cite{devlin2018bert}, with simple classification layers on top for pretraining. For downstream tasks, \ours \ can be used as a text-table encoder and easily integrated with any existing state-of-the-art model. {We conduct extensive experiments and show that:} 

(1) Combined with state-of-the-art text-to-SQL model RAT-SQL \cite{rat-sql}, using \ours \ as encoder significantly outperforms directly adopting pretrained BERT\textsubscript{LARGE} (RAT-SQL's default encoder) \add{and performs on par with other text-table pretraining models like GRAPPA \cite{yu2020grappa}} on the widely used Spider benchmark. \nop{To better demonstrate the superiority of our model in mapping between text and database schema, we also test the performance of RAT-SQL without the value linking component. This assumes the database content is not exposed to the semantic parser and prevents it from simply matching phrases in the utterance with the values stored in the column.}
% While Spider has covered many challenges in cross-domain text-to-SQL, it still has some limitations. For example, Spider includes many NL utterances that closely match their paired SQL query, a model can simply using string matching for column selection rather than really understand the utterance and database schema. To solve this, \citeauthor{suhr2020exploring} propose a new setting that includes 8 other single-domain text-to-SQL datasets for evaluation. However, some of the datasets contain many unique query structure and dataset conventions, making the generalization impossible without some in-domain training examples for adaptation. In light of this, we create a new evaluation set from the original Spider development set by removing the explicit mention of column names in the utterance. This allows us to better evaluate the generalization ability of the model and give a more fair assessment of how well the joint representation captures the association between the text and database schema.

(2) \add{On more realistic evaluation settings, including \newdata and the \citet{suhr2020exploring} datasets, our method outperforms all baselines.\nop{We also notice greater improvements using \ours\ under this setting.} This demonstrates the superiority of our \nop{model}{pretraining framework} in solving the text-table alignment challenge, and its usefulness in practice.}

(3) \ours \ also helps reduce the need for large amount of costly supervised training data. {We experiment with the WikiSQL benchmark \cite{zhongwikisql2017} by limiting training data size, and show that our pretraining method can boost the model performance by a large margin and consistently outperforms existing pretraining methods.} \nop{Experiment results on the WikiSQL benchmark show that using our pretrained model can boost the model performance by a large margin when only limited training data is available and brings consistent improvement over the baseline using the same amount of training data.}

\section{Related Work}
\noindent\textbf{Cross-Database Text-to-SQL.}
Remarkable progress has been made in text-to-SQL over the past few years. With sufficient in-domain training data, existing models already achieve over
80\% exact matching accuracy \cite{finegan2018improving, wang2018robust}\nop{the citations are a bit old; give more recent ones where 90\% can be found?} on single-domain benchmarks like ATIS \cite{hemphill1990atis,dahl1994expandingatis} and GeoQuery \cite{zelle1996geo}. However, annotating NL questions with SQL queries is expensive making it cost-prohibitive to collect training examples for all possible databases. A model that can generalize across domains and databases is desired. In light of this, \citet{yu2018spider} present Spider, a cross-database text-to-SQL benchmark that trains and evaluates a system using different databases. More recently, \citet{suhr2020exploring} provide a holistic analysis of the challenges introduced in cross-database text-to-SQL and propose to include single-domain datasets in evaluation. Their study uncovers the limitations of current text-to-SQL models, and demonstrates the need for models that can better handle the generalization challenges.

\nop{\noindent\textbf{Pretraining for Unstructured Text.}
Language model (LM) pretraining has shown promising results on improving the generalization ability of natural language processing (NLP) models on many tasks \cite{devlin2018bert, peters2018deep, liu2019roberta, lewis2019bart, guu2020realm}. Pretrained LMs like BERT \cite{devlin2018bert} can learn complex characteristics of word use across contexts from large-scale text corpus and apply the learned knowledge to downstream tasks. Aside from language modeling, multi-task learning (MTL) is another popular technique to pretrain a good text representation \cite{liu2015representation, liu2019multi}. \citet{liu2019multi} demonstrate that MTL provides an effective way of leveraging data from
many related tasks and is complementary to LM pretraining. However, as these models are usually pretrained with only unstructured text, they do not have knowledge about encoding structured data like database schema, or more importantly how to align them with the NL utterances.}

\noindent\textbf{Pretraining for Text-Table Data.}
Inspired by the success of pretrained language models, some recent work has tried to apply similar pretraining objectives to text-table data. TaBERT \cite{yin2020tabert} and TAPAS \cite{herzig2020tapas} jointly learn text-table representations by leveraging a large amount of web tables and their textual context. They flatten the tables and use special embeddings to model the structure information. {A masked language model (MLM) objective is then used to predict the masked tokens in the text-table data. MLM is good at modeling the contextualized semantic representations of a token, but is weak at capturing the alignment between a pair of sequences (e.g., text-table).}\nop{A masked language model (MLM) objective is then used to learn knowledge from the text-table data. However, MLM is originally designed for modeling\hs{\st{single}} NL sentences, making it weak at capturing the association between a pair of sequences, like the alignment between text and tables.} More recently, GRAPPA \cite{yu2020grappa} explores a different direction for pretraining which shares some similarity with existing work on data augmentation for semantic parsing. GRAPPA first constructs synthetic question-SQL pairs \add{using templates (a synchronous context free grammar) induced}\nop{via a synchronous context free grammar induced} from existing text-to-SQL datasets, \nop{I feel the following sentence is hard for reviewers to get without reading GRAPPA} a SQL semantic prediction objective is then used to learn compositional inductive bias from the synthetic data. \add{However, as the synthetic data is generated using templates, and the column names and values are directly filled in the questions, it has the same problem as existing text-to-SQL datasets that eases the text-table alignment challenge. In constrast, \ours\ aims to directly learn the text-table alignment knowledge from parallel text-table corpora via structure-grounded pretraining objectives.\nop{, which will greatly benefit the downstream text-to-SQL task.} We also note that existing pretraining methods and \ours\ can be complementary and combined together in the future.} \nop{However, the strength of GRAPPA mainly comes from using the synthetic data. There still lack models that can fully utilize the alignment knowledge naturally present in parallel text-table corpora, which is the aim of this work.}\nop{How about changing the last sentence to: Compared with these methods, our \ours aims to directly learn the text-table alignment knowledge via carefully-designed objectives in pretraining, which will greatly benefit the downstream text-to-SQL task. We also note that existing pretraining methods and \ours can be complementary and combined together in the future.}

\noindent\textbf{Structure Grounding in Text-to-SQL.}
{Structure grounding has been proven to be crucial for text-to-SQL, where a model needs to correctly identify column and value mentions in an NL utterance and link them to the given database schema \cite{guo2019irnet, bogin2019global,rat-sql,lei2020re}. Most existing text-to-SQL systems have specially designed components for structure grounding, which is also referred to as schema linking. {For example, \citet{guo2019irnet, yu2018typesql} explore using simple heuristics like string matching for schema linking, and use the linking results as direct hints to their systems.} However, such heuristics may not generalize well in real world scenarios where there are varied ways to refer to a column, which usually differ from the original column name. More recently, \citet{Shi:Zhao:Boyd-Graber:Daume-III:Lee-2020} and \citet{lei2020re} take a step forward and {manually} annotate WikiTableQuestions \cite{pasupat2015compositional} and Spider with fine-grained alignment labels for supervised training\nop{of structure grounding} {(together with the text-to-SQL objective)}, which brings significant improvements. The main drawback of these models is that\nop{they still only learn} they are limited to learn the alignment knowledge from a relatively small training corpus, and cannot generalize well in a cross-domain setting. Moreover, SQL annotations and fine-grained alignment labels \nop{why are their fine-grained alignment labels costly? Can they not use the SQL annotations to obtain?} are both expensive to get {manually}. In contrast, this paper aims to {re-purpose an existing parallel text-table corpus for pretraining models to learn structure grounding}, where we generate alignment labels\nop{in a semi-supervised fashion} at large scale with low or no cost.}

\section{Structure-Grounded Pretraining}
\nop{How about changing the title to: Structure-Grounded Pretraining? I feel that is the key; weak supervision and mutl-task are how we realized it, but might not sound novel.}
\subsection{Motivation}
One of the critical generalization challenges in cross-database text-to-SQL is text-table alignment, i.e., a model needs to understand NL utterances and database schemas unseen in training, including value mentions and novel columns, and to correctly map between them. Similar generalization challenges have been studied for a long time in the NLP field. Recently, pretrained language models \cite{devlin2018bert, liu2019roberta, lewis2019bart} have achieved great success in tackling the challenges by learning\nop{syntactic and semantic} contextualized representations of words\nop{ across contexts} from a large text corpus. \add{Inspired by this, in this work we aim to develop a pretraining method that can directly learn the text-table alignment knowledge from a large parallel text-table corpus.\nop{is the last sentence precise? I feel it seems to imply we also have MLM as objective. how about: in this work we aim to develop a pretraining method which can directly learn the text-table alignment knowledge.}}

Unlike previous \add{text-table pretraining} works \cite{herzig2020tapas,yin2020tabert} that optimize unsupervised objectives like MLM during pretraining, {we carefully design three structure-grounded tasks}\nop{aim to directly capture the text-table alignment knowledge via three structure-grounded tasks}: column grounding, value grounding and column-value mapping.\nop{Such multi-task objective benefit from two advantages: (1) Task-specific objective is easier to optimize compared with MLM, saving great amount of training time and computing resource. (2) MLM is originally designed for modeling single NL sentence, and is weak at capturing the association between pair of sequences. \citeauthor{herzig2020tapas} also experiment with a sentence-table entailment objective but do not find it helpful for downstream task.}\nop{The following two sentences are the key, IMHO. Move up or highlight or reduce the sentences earlier in this paragraph} {These tasks are related to text-to-SQL and can directly capture the text-table alignment during pretraining.}
\nop{Our tasks are designed for finding the association\nop{correlation} between the unstructured text and structured table schema and are closely related to text-to-SQL.}As a result, the learned alignment knowledge can be effectively transferred to the downstream task and improve the final performance.

\subsection{Pretraining Objectives}
We use the same model architecture as BERT, and add simple classification layers on top for the three structure-grounded tasks. For downstream tasks, our model can be easily integrated into existing models as text-table encoder. Following previous work \cite{hwang2019sqlova,rat-sql,guo2019irnet}, we linearize the input by concatenating the NL utterance and column headers, using \texttt{\textless sep\textgreater} token as a separator.

Formally, given a pair of NL utterance $\left\{x_i\right\}$ and table with a list of column headers (\add{in case there are multiple tables like in databases, we concatenate all the column names together}) $\left\{c_j\right\}$,\nop{if there are multiple tables in a database, we concatenate column names of all tables, right?} we first obtain the contextualized representation $\mathbf{x}_i$ of each token in the utterance and $\mathbf{c}_j$ for each column using the last layer output of the BERT encoder. Here each column header $c_j$ may contain multiple tokens $c_{j,0},\dots,c_{j,|c_j|}$. We obtain a single vector representation for each column using column pooling. More specifically, we take the output of the first and last token of the header, and calculate the column representation as $\mathbf{c}_j = (\mathbf{c}_{j,0}+\mathbf{c}_{j,|c_j|})/2$. $\{\mathbf{x}_i\}$ and $\{\mathbf{c}_j\}$ are then used to compute losses for the three tasks. An overview of our model architecture and pretraining objectives are shown in \cref{fig:objective}.\nop{maybe we could refer to table 2 earlier, e.g., in Intro, or move Figure 2 to this page.}
\begin{table*}[t]
    \centering
    \resizebox{0.8\linewidth}{!}{
    \begin{tabular}{lccc}
    \toprule
        Dataset & \makecell{\# Examples\nop{\\\cite{suhr2020exploring}}}& \makecell{Exec Acc\\\cite{suhr2020exploring}}\nop{& \makecell{\# Examples\\Our Transformation}}& \makecell{\% Col Mentioned}\\
        \midrule
         ATIS \cite{hemphill1990atis,dahl1994expandingatis}&289 (486)& 0.8&\nop{283-&}0.0\\
         Restaurants \cite{tang2000restaurants}&\hspace{5pt}27 (378)& 3.7&\nop{39-&}0.0\\
         Academic\cite{li2014academic}& 180 (196)& 8.2&\nop{167-&}11.4\nop{5.6}\\
         Yelp\cite{yaghmazadeh2017sqlizer}&\hspace{5pt}54 (128)&19.8&\nop{68-&}8.0\nop{7.7}\\
         Scholar\cite{iyer2017learning}&394 (599)&0.5&\nop{399-&}0.0\\
         Advising\cite{finegan2018improving}&\hspace{5pt}309 (2858)&2.3&\nop{304-&}0.3\nop{1.0}\\
         IMDB\cite{yaghmazadeh2017sqlizer}&107 (131) &24.6&\nop{111&}1.0\nop{0.9}\\
         GeoQuery\cite{zelle1996geo}& 532 (598) & 41.6&\nop{525&}3.9\\
         \midrule
         Spider \cite{yu2018spider}&1034&69.0&\nop{1034&}39.2\\
         \newdata&508 &-&\nop{508&}1.8\\
        \bottomrule
    \end{tabular}}
    \caption{Statistic of the datasets used in this work. Here we show the number of examples for evaluation after filtering (sizes of the original datasets before any filtering are shown in parentheses), and the execution accuracy reported in \citet{suhr2020exploring}. For the detailed filtering process of \citet{suhr2020exploring}, please check the original paper or \cref{sec:data_filter}.\nop{We follow the descriptions in \citet{suhr2020exploring} \add{to filter IMDB and GeoQuery, and transform them to a uniform format as Spider}, statistics of the resulted datasets are shown in our transformation above.} \% Col Mentioned\footnotemark measures the proportion of examples in the evaluation set where all columns compared against entities in the gold query are explicitly mentioned in the NL utterance. \nop{For more details on preparing the datasets for evaluation, please check \cref{sec:data_details}.} \nop{in your Teams pptx, the numbers for \% col mentioned are all higher (Spider is 72.4\%, Academic is 8.2\%, etc.). I think you were doing stemming to compute this number. I think this makes it an even stronger case if you use these stemmed numbers (GeoQuery does become quite large (32.4\%) but the rest stay small, so I think the tradeoff is worth it).} \nop{The numbers in the slides are from \citet{suhr2020exploring}. The only way I can match their number is to consider all examples that does not contain where columns as positive (there is no column to mention so all column is mentioned). Here I am using the ratio of (examples all columns are mentioned)/(examples that contain at least one column)}}
    \vspace{-10pt}
    \label{tab:data}
\end{table*}

\noindent\textbf{Column grounding.} An important task in text-to-SQL is to identify grounded columns from the schema and use them for the generated SQL query.\nop{SELECT? I don't think you need to list the clause types} With a parallel text-table corpus, this is similar to selecting the columns that are mentioned in the associated NL sentence. This task requires a model to understand the semantic meaning of a column based on its header alone\nop{(without table cells under this column)}, and to infer its relation with the NL sentence based on the contextualized representations. We formulate it as a binary classification task. For each column $c_j$, we use a one-layer feed forward network $f(\cdot)$ to get prediction $p^c_j=f(\mathbf{c}_j)$ of whether $c_j$ is mentioned in the sentence or not. The column grounding loss $\mathcal{L}_c$ is then calculated using the binary cross entropy loss w.r.t. ground truth labels $y^c_j\in \{0,1\}$. Note this task requires the model to identify the meaning of a column without access to any of its values. Hence, it is suitable for the typical text-to-SQL setting where the model only has access to the database schema.

\noindent\textbf{Value grounding.} For clauses like \texttt{WHERE} and \texttt{HAVING}, to generate an executable SQL query, a model also needs to extract the value to be compared with the grounded column from the NL utterance. This can be transformed to the task of finding cell mentions in the NL sentence with a parallel text-table corpus. \nop{The model needs to first identify phrases that are entities and then keep only those related to the associated table.} Since the contents of the table is not available\nop{for text-to-SQL}, it is necessary for the model to infer the possible value \nop{types}\hs{mentions} based on {NL utterance and} the table schema only\nop{and use it to filter the values}. Similarly to column grounding, we also view this as a classification task. For each token $x_i$, we get prediction of $x_i$ being part of a grounded value as $p^v_i=f(\mathbf{x}_i)$. The value grounding loss $\mathcal{L}_v$ is then calculated using the binary cross entropy loss w.r.t. ground truth labels $y^v_i\in \{0,1\}$. \nop{access to table values? Sounds like it does have access but math says no?}

\noindent\textbf{Column-Value mapping.} As there may be multiple columns and values used in the SQL query, a text-to-SQL model also needs to correctly map the grounded columns and values. This is used to further strengthen the model's ability to capture the correlation between the two input sequences by learning to align the columns and values\nop{ grounded in previous two tasks}. We formulate this as a matching task between the tokens in the NL sentence and the columns. For every \add{grounded} token $x_i$ \add{(i.e., $y^v_i=1$)}\nop{does $x_i$ have to be a value token? what about the `department' token in previous example?}, we pair it with each column $c_j$ and calculate the probability of $x_i$ matching $c_j$ as $p^{cv}_{i,j}=f([\mathbf{x}_i,\mathbf{c}_j])$. Here $[\cdot,\cdot]$ is the vector concatenation operation. We then apply a softmax layer over the predictions for each token $p^{cv}_{i} = \{p^{cv}_{i,j}\}_{j=1}^{|c|}$, and the final column-value mapping loss $\mathcal{L}_{cv}$ is then calculated as $\mathcal{L}_{cv}=\mathrm{CrossEntropy}\left(\mathrm{softmax}\left(p^{cv}_{i}\right),y^{cv}_i\right)$, where $y^{cv}_i \in \{0,1\}^{|c|}$ is the ground truth label.

The final loss $\mathcal{L}$ \hs{for pretraining} is the sum of all three losses. We experimented with different weights for each term, but did not observe significant improvement on the results. Hence we only report results with equally weighted losses.
\nop{Did you try having only the column-value mapping loss?}
\vspace{-5pt}
\begin{equation}
    \mathcal{L} = \mathcal{L}_c + \mathcal{L}_v + \mathcal{L}_{cv}
\end{equation}

\subsection{Obtaining Pretraining Data via Weak Supervision}
\footnotetext{Unlike \citet{suhr2020exploring}, here we do not consider examples where there is no column compared against entity.}
We obtain ground truth labels $y^c_j$, $y^v_i$ and $y^{cv}_i$ from a parallel text-table corpus based on a simple intuition: given a column in the table, if any of its cell values can be matched to a phrase in the sentence, this column is likely mentioned in the sentence, and the matched phrase is the value aligned with the column. To ensure high quality text-table alignment information in the pretraining corpus, unlike previous work \cite{herzig2020tapas,yin2020tabert} that use loosely connected web tables and their surrounding text, here we leverage an existing large-scale table-to-text generation dataset ToTTo \cite{parikh2020totto}. ToTTo contains  120,761 NL descriptions and corresponding web tables automatically collected from Wikipedia using heuristics. Additionally, it provides cell level annotation that highlights cells mentioned in the description and revised version of the NL descriptions with irrelevant or ambiguous phrases removed. 

We experiment with two pretraining settings, with or without human assistance. In the \textit{human assisted setting}, we use the cell annotations along with the revised description \add{to infer the ground truth labels}. More specifically, we first label all the columns $c_j$ that contain at least one highlighted cell as positive ($y^c_j=1$). We then iterate through all the values of the highlighted cells and match them with the NL description \add{ via exact string matching} to extract value mentions. If a phrase is matched to a highlighted cell, we select all the tokens $x_i$ in that phrase and align them with the corresponding columns $c_j$ ($y^v_i=1$, $y^{cv}_{i,j}=1$). In the \textit{automatic setting}, we use only the tables and the raw sentences, and obtain cell annotations by comparing each cell with the NL sentence using exact string matching. {Note that in both settings, the cell values are used only for \hs{preparing supervision for the pretraining objectives}\nop{pretraining data construction}, not as inputs to the pretraining model.} \nop{For more details on preparing the pretraining data, please see \cref{sec:pretrain_data_detail}.}

To make the pretraining more effective and to achieve a better generalization performance, we also incorporate two data augmentation techniques. First, since the original parallel corpus only contains one table for each training example, we randomly sample $K_{neg}$ tables as negative samples and append {their column names} to the input sequence. This simulates a database with multiple tables and potentially hundreds of columns, which is common in text-to-SQL. Second, we randomly replace the matched phrases in the NL sentences with values of cells from the same column \add{(the labels are kept the same)}.\nop{Do we want to talk more about this data augmentation? e.g., for each training example, you create 2 additional examples? Do we have ablation studies to show how effective this is?} This way we can better leverage the contents of the table \hs{during pretraining} and improve the model's generalization ability by exposing it to more cell values.
\nop{Can we come up with names for these settings?}
\section{Creating a More Realistic Evaluation Set}
\label{sec:new_eva}
As one of the first datasets to study cross-database text-to-SQL, Spider has been a widely used benchmark in assessing a model's ability to generalize to unseen programs and databases. However, as pointed out by \citet{suhr2020exploring}, Spider eases the task by using utterances that closely match their paired SQL queries, for example by explicitly mentioning the column names in the question, while in practice NL references to columns usually differ from the original column name. To alleviate this problem, \citet{suhr2020exploring} propose to train the model with cross-domain dataset like Spider, and add another eight single-domain datasets like ATIS \cite{hemphill1990atis,dahl1994expandingatis} and GeoQuery \cite{zelle1996geo} for evaluation. However, some of the datasets differ a lot from Spider, introducing many novel query structures and dataset conventions.\footnote{\add{Some of the datasets contain operators that are not covered by Spider grammar or novel query structure like self join that does not exist in the training corpus.}} As we can see from \cref{tab:data}, their model \cite{suhr2020exploring} has very poor performance in some datasets. In light of this, we present a new realistic and challenging evaluation set based on Spider. We first select a complex subset from the Spider dev set where there are columns compared against values or used in clauses like \texttt{ORDER BY}. We then manually modify the NL questions in the subset \add{ourselves} to remove or paraphrase explicit mentions of columns names, except for the columns in \texttt{SELECT} clauses, while keeping the SQL queries unchanged. Some examples are shown in \cref{tab:new_eva}. This way we do not introduce extra challenges like adapting to new query structures but make it possible to fairly assess the model's capability in aligning text and tables. \nop{We also include two datasets from \citet{suhr2020exploring}, IMDB \cite{yaghmazadeh2017sqlizer} and GeoQuery, on which existing models can achieve fair performance without finetuning with in-domain data. We follow the same filtering process as \citet{suhr2020exploring} and transform the datasets to the same format as Spider. Examples that are not compatible with Spider grammar are removed. }\add{To make a more comprehensive comparison, we will also report results on the original \citet{suhr2020exploring} datasets.} \nop{Huan: I wonder since we are also reporting on their original datasets, why do we still transform IMDB and GeoQuery and report on them separately in Table 4 and 5?}

\section{Experiments}
\subsection{Benchmarks and Base Models}
\begin{table}[]
    \centering
    \resizebox{0.9\linewidth}{!}{
    \begin{tabularx}{1.05\linewidth}{ll}
        \toprule
        \small Example & \small Type \\
        \midrule
        \makecell[l]{\small Show name, country, age for all singers \textcolor{red}{\sout{ordered}}\\ \small \textcolor{red}{\sout{ by \textit{age}}} from the oldest to the youngest.}&\small Remove\\
        \midrule
        \makecell[l]{\small Find the number of concerts happened in the\\ \small stadium \textcolor{red}{\sout{with the highest \textit{capacity}}} \textcolor{blue}{that can}\\ \textcolor{blue}{\small accommodate the most people}.}&\multirow{2}{*}{\small paraphrase}\\
        \makecell[l]{\small How many pets \textcolor{red}{\sout{have a greater \textit{weight} than 10}}\\ \small \textcolor{blue}{are over 10 lbs}?}&\\
        \bottomrule
    \end{tabularx}}
    \caption{Examples of how we create \newdata from Spider. Phrases shown in italic exactly match with column names.}
    \label{tab:new_eva}
    \vspace{-15pt}
\end{table}

\begin{table*}[ht]
    \centering
    \resizebox{\linewidth}{!}{
    \begin{tabular}{llccccccccc}
    \toprule
        &Models&Spider-Realistic&ATIS&GeoQuery&Restaurants&Academic&IMDB&Yelp&Scholar&Advising\\
        \midrule
        &\# Examples&508&289&532&27&180&107&54&394&309 \\
        \midrule
        \multirow{5}{*}{\rotatebox[origin=c]{90}{Schema Only}}&\citet{suhr2020exploring}&-&0.8 (0.5)& 41.6 (35.6)&3.7 (3.7)&8.2 (6.1)&24.6 (24.3)&\textbf{19.8 (16.7)}&0.5 (0.4)&2.3 (1.2)\\
        &RAT-SQL \textit{w/o} value linking &&&& && & \\
        &\hspace{10pt}\textit{w.} BERT\textsubscript{LARGE} &52.4 $\pm$ 0.7 (46.9)&2.1 $\pm$ 0.6 &41.2 $\pm$ 11.6&0.0 $\pm$ 0.0&5.9 $\pm$ 2.1&26.5 $\pm$ 5.0&12.3 $\pm$ 1.7&0.8 $\pm$ 0.4&1.6 $\pm$ 0.4 \\
        &\hspace{10pt}\textit{w.} \ours \ (Human Assisted) &57.8 $\pm$ 0.6 (53.3)&2.2 $\pm$ 0.2 &45.5 $\pm$ 1.8&11.1 $\pm$ 9.1&\textbf{14.8 $\pm$ 5.0}&\textbf{37.1 $\pm$ 1.8}&15.4 $\pm$ 0.9&4.3 $\pm$ 1.7&\textbf{2.2 $\pm$ 0.4}\\
        &\hspace{10pt}\textit{w.} \ours \ (Automatic) &\textbf{60.3 $\pm$ 0.7 (54.9)}&\textbf{2.2 $\pm$ 0.2} &\textbf{50.9 $\pm$ 4.0}&\textbf{40.7 $\pm$ 5.2}&12.4 $\pm$ 1.9&35.5 $\pm$ 2.0&13.0 $\pm$ 2.6&\textbf{5.4 $\pm$ 0.7}&1.0 $\pm$ 0.3 \\
        \midrule
         \multirow{5}{*}{\rotatebox[origin=c]{90}{Content Used}}&RAT-SQL & && && & \\
        &\hspace{10pt}\textit{w.} BERT\textsubscript{LARGE} &62.1 $\pm$ 1.3 (58.1)&2.3 $\pm$ 0.2 &47.3 $\pm$ 3.7&37.0 $\pm$ 18.9&15.6 $\pm$ 2.0&21.8 $\pm$ 1.6&16.0 $\pm$ 3.1&3.4 $\pm$ 1.4&6.4 $\pm$ 2.3 \\
        &\hspace{10pt}\textit{w.} GRAPPA &- (59.3) && && & \\
        &\hspace{10pt}\textit{w.} \ours \ (Human Assisted) &\textbf{65.7 $\pm$ 0.7 (62.2)}&\textbf{5.5 $\pm$ 1.1} &\textbf{59.5 $\pm$ 3.2}&40.7 $\pm$ 13.9&18.7 $\pm$ 2.1&26.8 $\pm$ 2.9&\textbf{21.6 $\pm$ 2.3}&\textbf{6.3 $\pm$ 1.8}&\textbf{6.9 $\pm$ 0.6} \\
        &\hspace{10pt}\textit{w.} \ours \ (Automatic) &65.3 $\pm$ 0.7 (62.2)&2.8 $\pm$ 0.7 &57.5 $\pm$ 0.2&\textbf{44.4 $\pm$ 32.7}&\textbf{20.2 $\pm$ 1.6}&\textbf{30.2 $\pm$ 5.8}&18.5 $\pm$ 1.5&6.1 $\pm$ 0.5&5.2 $\pm$ 0.5 \\
        \bottomrule
        
    \end{tabular}}
    \vspace{-5pt}
    \caption{Execution accuracy on the more realistic evaluation sets including \newdata and the \citet{suhr2020exploring} evaluation sets. For Spider-Realistic, we also show exact match accuracy in parentheses. For the \citet{suhr2020exploring} evaluation sets, we show results for the filtered set where examples with query returning empty set are excluded. \citet{suhr2020exploring} uses the WikiSQL dataset as additional training data, and we also show their results with only the Spider training data in parentheses.}
    \vspace{-5pt}
    \label{tab:suhr_original}
\end{table*}
\begin{table}[]
    \centering
    \resizebox{\linewidth}{!}{
    \begin{tabular}{clccc}
    \toprule
        &Models & Exact & Exec & Exact (Test)\\
        \midrule
        \multirow{8}{*}{\rotatebox[origin=c]{90}{Schema Only}}&EditSQL \cite{zhang2019editing} \textit{w.} BERT &57.6 &-&53.4 \\
        &IRNET \cite{guo2019irnet} \textit{w.} BERT & 61.9& -&54.7 \\
        &RYANSQL \cite{choi2020ryansql} \textit{w.} BERT &\textbf{70.6} & -&60.6 \\
        &\citet{suhr2020exploring} \textit{w.} BERT\textsubscript{LARGE+} &65.0&69.0&-\\
        &RAT-SQL \cite{rat-sql} \textit{w/o} value linking & & &\\
        &\hspace{10pt}\textit{w.} BERT\textsubscript{LARGE} &67.0 $\pm$ 0.6&69.8 $\pm$ 0.3&-\\
        &\hspace{10pt}\textit{w.} \ours \ (Human Assisted) &70.5 $\pm$ 0.6&73.3 $\pm$ 0.4&\textbf{67.4}\\
        &\hspace{10pt}\textit{w.} \ours \ (Automatic) &69.8 $\pm$ 0.3&\textbf{74.2 $\pm$ 0.8}&-\\
        \midrule
         \multirow{7}{*}{\rotatebox[origin=c]{90}{Content Used}}&Global-GNN \cite{bogin2019global} & 52.7 & -&47.4 \\
         &TranX \textit{w.} TaBERT \cite{yin2020tabert} & 64.5 &-&- \\
         &RAT-SQL \textit & & &\\
        &\hspace{10pt}\textit{w.} BERT\textsubscript{LARGE} &69.8 $\pm$ 0.8&72.3 $\pm$ 0.6&-\\
        &\hspace{10pt}\textit{w.} GRAPPA \cite{yu2020grappa} &\textbf{73.4} & -&\textbf{69.6}\\
        &\hspace{10pt}\textit{w.} \ours \ (Human Assisted) &72.7 $\pm$ 0.7&\textbf{75.5 $\pm$ 0.8}&68.4\\
        &\hspace{10pt}\textit{w.} \ours \ (Automatic) &72.6 $\pm$ 0.1&74.9 $\pm$ 0.1&-\\
        \bottomrule
    \end{tabular}}
    \vspace{-5pt}
    \caption{Results on Spider. The top half shows models using only database schema, the bottom half shows models using the database content. We train our model three times with different random seeds and report the mean and standard deviation here.}
    \vspace{-5pt}
    \label{tab:spider}
\end{table}

% \begin{table}[t]
%     \centering
%     \resizebox{\linewidth}{!}{
%     \begin{tabular}{lcccccc}
%     \toprule
%         \multirow{2}{*}{Models}&\multicolumn{2}{c}{\newdata}&\multicolumn{2}{c}{IMDB}&\multicolumn{2}{c}{Geo}\\
%         & Exact & Exec& Exact & Exec& Exact & Exec\\
%         \midrule
%         RAT-SQL \textit{w/o} value linking &&& && & \\
%         \hspace{10pt}\textit{w.} BERT\textsubscript{LARGE} &46.9 &52.4&23.7 &28.8&12.8 &40.3 \\
%         \hspace{10pt}\textit{w.} \ours \ (Human Assisted) &53.3 &57.8&\textbf{32.7} &\textbf{38.1}&14.8 &44.3 \\
%         \hspace{10pt}\textit{w.} \ours \ (Automatic) &\textbf{54.9} &\textbf{60.3}&31.5 &36.6&\textbf{17.9} &\textbf{48.7} \\
%         \midrule
%          RAT-SQL & && && & \\
%         \hspace{10pt}\textit{w.} BERT\textsubscript{LARGE} &58.1 &62.1&15.9 &24.9&20.4 &46.7 \\
%         \hspace{10pt}\textit{w.} GRAPPA &59.3 &-&- &-&- &- \\
%         \hspace{10pt}\textit{w.} \ours \ (Human Assisted) &\textbf{62.2} &\textbf{65.7}&20.4 &28.8&23.6 &\textbf{57.6} \\
%         \hspace{10pt}\textit{w.} \ours \ (Automatic) &62.2 &65.3&\textbf{26.4} &\textbf{32.4}&\textbf{24.1} &55.9 \\
%         \bottomrule
%     \end{tabular}}
%     \vspace{-5pt}
%     \caption{Results on the more realistic evaluation sets. Here we report the average performance.}
%     \vspace{-15pt}
%     \label{tab:spider_modified}
% \end{table}
\begin{table}[t]
    \centering
    \resizebox{0.75\linewidth}{!}{
    \begin{tabular}{lcc}
    \toprule
        Models & ACC\textsubscript{lf}& ACC\textsubscript{ex}\\
        \midrule
        HydraNet \cite{lyu2020hybrid} &\textbf{83.8} &\textbf{89.2} \\
        X-SQL \cite{he2019xsql}  &83.3 &88.7 \\
        SQLova \cite{hwang2019sqlova} &&\\
        \hspace{10pt}\textit{w.} BERT\textsubscript{LARGE} &82.1 &87.3 \\
        \hspace{10pt}\textit{w.} TaBERT &\textbf{82.5} &\textbf{87.9} \\
        \hspace{10pt}\textit{w.} \ours \ (Human Assisted) &82.1 &87.5 \\
        \hspace{10pt}\textit{w.} \ours \ (Automatic) &82.4&87.8\\
        \midrule
        SQLova (5\%) & &\\
        \hspace{10pt}\textit{w.} BERT\textsubscript{LARGE} &70.7 &77.0 \\
        \hspace{10pt}\textit{w.} TaBERT &71.5 &78.0 \\
        \hspace{10pt}\textit{w.} \ours \ (Human Assisted) &\textbf{75.6} &\textbf{81.6} \\
        \hspace{10pt}\textit{w.} \ours \ (Automatic) &75.6 &81.4 \\
         \bottomrule
    \end{tabular}}
    \vspace{-5pt}
    \caption{Performance on WikiSQL. Here we show logical form accuracy and execution accuracy on the test set. (5\%) means random sampling 5\% of original training data for training.}
    \vspace{-5pt}
    \label{tab:wikisql}
\end{table}
\noindent\textbf{Spider and the realistic evaluation sets.} Spider \cite{yu2018spider} is a complex cross-database text-to-SQL dataset.  It contains 10k complex question-query pairs grounded on 200 databases where multiple tables are joined via foreign keys. In addition, we create a new realistic evaluation set \newdata as described in \cref{sec:new_eva}. \nop{We also include IMDB and GeoQuery, two of the eight single-domain datasets used in \citet{suhr2020exploring}, for a more comprehensive comparison.}\add{We also \nop{transform IMDB and GeoQuery to the same format as Spider, and }include the original \citet{suhr2020exploring} datasets, for a more comprehensive comparison.} For the \hs{base}\nop{baseline} model, we use RAT-SQL \cite{rat-sql} which is the state-of-the-art model according to the official leaderboard as of the submission time. To generate executable SQL queries, we modify the pointer generator in RAT-SQL to enable it to copy values from the question. We use the same trained model for evaluation on the Spider dev set and the realistic evaluation sets. \citet{yu2018spider} includes some single-domain text-to-SQL datasets like GeoQuery as extra training data for Spider. Following \citet{suhr2020exploring}, we train the model with only the original Spider data, and discard additional training data used by some previous works \hs{like \citet{yu2018spider}}. We use both the set match accuracy (exact match) from the official Spider evaluation script and execution accuracy\footnote{We execute case insensitive SQL queries, and compare the returned table.} for evaluation on Spider and Spider-Realistic.\nop{ and the transformed IMDB and GeoQuery.} On the \citet{suhr2020exploring} datasets, we use the official evaluation script\footnote{\url{https://github.com/google-research/language/tree/master/language/xsp}} released by the authors and report execution accuracy.

\noindent\textbf{WikiSQL.} WikiSQL \cite{zhongwikisql2017} is a large-scale text-to-SQL dataset consists of over 80k question-query pairs grounded on over 30k Wikipedia tables. Although existing models are already reaching the upper-bound performance on this dataset \cite{hwang2019sqlova,yavuz2018takes}, mainly because of the simplicity of the SQL queries and large amount of data available for training, previous works have also used this dataset to demonstrate the model's generalization ability with limited training data \cite{yu2020grappa, yao2020imitation}. For the \hs{base} model, we use SQLova \cite{hwang2019sqlova} without execution-guided decoding. Following the official leaderboard, we report both logical form accuracy and execution accuracy. \nop{To better demonstrate how our pretraining objectives help with the text-to-SQL task, we also show break-down accuracy for several subtasks.}

% For more implementation details, please see \cref{sec:training_details}.
\subsection{Training Details}
\label{sec:training_details}
For all experiments, we use the BERT implementation from Huggingface \cite{Wolf2019HuggingFacesTS} and the pretrained BERT\textsubscript{LARGE} model from Google \footnote{We use the BERT-Large, Uncased (Whole Word Masking) model from \url{https://storage.googleapis.com/bert_models/2019_05_30/wwm_uncased_L-24_H-1024_A-16.zip}}. For pretraining, we use Adam optimizer \cite{kingma2014adam} with a initial learning rate of 2e-5 and batch size of 48. In both settings, we use $K_{neg}=1$ and pretrains our model for 5 epochs. We use 4 V100 GPUs for pretraining, which takes less than 4 hours.

For Spider and the realistic evaluation sets, we use the official implementation of RAT-SQL \footnote{\url{https://github.com/microsoft/rat-sql}} and modify it to generate executable SQL queries. We follow the original settings and do hyperparameter search for learning rate (3e-4, 7.44e-4) and warmup step (5k, 10k). We use the same polynomial learning rate scheduler with warmup and train for 40,000 steps with batch size of 24. The learning rate for the pretrained encoder (e.g. BERT) is 3e-6 and is frozen during warmup.

For WikiSQL, we use the official SQLova implementation \footnote{\url{https://github.com/naver/sqlova}}. We use the default setting with learning rate of 1e-3 for the main model and learning rate of 1e-5 for the pretrained encoder. We train the model for up to 50 epochs and select the best model using the dev set.
%on pretraining and finetuning the models
% \begin{figure*}
% \begin{subfigure}{.33\linewidth}
%   \centering
%   \includegraphics[width=\linewidth]{figures/wikisql_x.png}
%   \caption{Execution Accuracy}
%   \label{fig:x_change}
% \end{subfigure}%
% \begin{subfigure}{.33\linewidth}
%   \centering
%   \includegraphics[width=\linewidth]{figures/wikisql_wc.png}
%   \caption{Where Column Accuracy}
%   \label{fig:wc_change}
% \end{subfigure}
% \begin{subfigure}{.33\linewidth}
%   \centering
%   \includegraphics[width=\linewidth]{figures/wikisql_wv.png}
%   \caption{Where Value Accuracy}
%   \label{fig:wv_change}
% \end{subfigure}
% \caption{Model performance on the test set with different fractions of training data.}
% \label{fig:wikisql_change}
% \end{figure*}

% \begin{figure*}
% \begin{subfigure}{.33\linewidth}
%   \centering
%   \includegraphics[width=\linewidth]{figures/wikisql_x_train.png}
%   \caption{Execution Accuracy}
%   \label{fig:x_train}
% \end{subfigure}%
% \begin{subfigure}{.33\linewidth}
%   \centering
%   \includegraphics[width=\linewidth]{figures/wikisql_wc_train.png}
%   \caption{Where Column Accuracy}
%   \label{fig:wc_train}
% \end{subfigure}
% \begin{subfigure}{.33\linewidth}
%   \centering
%   \includegraphics[width=\linewidth]{figures/wikisql_wv_train.png}
%   \caption{Where Value Accuracy}
%   \label{fig:wv_train}
% \end{subfigure}
% \caption{Model performance on the development set during training with 5\% of training data.}
% \label{fig:wikisql_train}
% \end{figure*}
\subsection{Main Results}
\noindent\textbf{Spider.}
We first show results on Spider dev set in \cref{tab:spider}. The original Spider setting assumes only the schema information about the target database is known in both training and evaluation phase, as the content of the database may not be accessible to the system due to privacy concern. More recently, some works have tried to using the database content to help understand the columns and link with the NL utterance. Here we show results for both settings. In the first setting where only schema information is known, we disable the value-based linking module in RAT-SQL. As we can see from \cref{tab:spider}, replacing BERT\textsubscript{LARGE} with \ours \ consistently improves the model performance in both settings. Under the setting where content is available, using \ours \ achieves similar performance as GRAPPA and outperforms all other models. GRAPPA uses both synthetic data and larger text-table corpus for pretraining. However, it mainly learns inductive bias from the synthetic data while our model focuses on learning text-table association knowledge from the parallel text-table data. \add{In error analysis on the Spider dev set, we notice that our best model\footnote{RAT-SQL \textit{w.} \ours \ (Human Assisted)} corrects 76 out of 270 wrong predictions made by GRAPPA while GRAPPA corrects 80 out of 274 wrong predictions made by our model.} This demonstrates that the two pretraining techniques are complementary and we expect combining them can lead to further performance improvement. For results on different difficulty levels and components, please see \cref{sec:more_results_spider}.

\noindent\textbf{More realistic evaluation sets.}
Results on the realistic evaluation sets are summarized in \cref{tab:suhr_original}. Firstly, we notice the performance of all models drops significantly on Spider-Realistic, demonstrating that inferring columns without explicit hint is a challenging task and there is much room for improvement. \add{Secondly, using \ours\ brings consistent improvement over BERT\textsubscript{LARGE} in all realistic evaluation sets. In the \newdata\ set, using \ours\ also outperforms GRAPPA\footnote{We use the checkpoint provided by the author, which achieves 73.8\% exact match accuracy on the Spider dev set. Here we only evaluate on \newdata\ with exact match accuracy because their model does not generate values and includes IMDB and Geo as extra training data. } by 2.9\%. Under the original \citet{suhr2020exploring} setting, combining RAT-SQL with \ours\ significantly outperforms \citet{suhr2020exploring} in all datasets, \hs{despite that we do not include WikiSQL as additional training data as they did}\nop{even we do not include WikiSQL as additional training data}.} Thirdly, comparing results in \cref{tab:spider} with \cref{tab:suhr_original}, using \ours\ brings larger improvement over BERT\textsubscript{LARGE} in the more realistic evaluation sets. As shown in \cref{tab:data}, the original Spider dataset has a high column mention ratio, so the models can use exact match for column grounding without really understanding the utterance and database schema. The more realistic evaluation sets better simulate the real world scenario and contain much less such explicit clues, making the text-table alignment knowledge learned by \ours\ more valuable. For case studies on Spider-Realistic, please check \cref{sec:case}. \nop{Interestingly, we also notice that using value linking actually hurts the performance on IMDB. One possible explanation is that in IMDB value linking may provide many noisy links, e.g. a person's name can be linked to multiple columns including actor\_name, writer\_name, character\_name and even film\_title, this may confuse the model and leads to wrong prediction.}

\begin{figure}[t]
  \centering
  \includegraphics[width=0.75\linewidth]{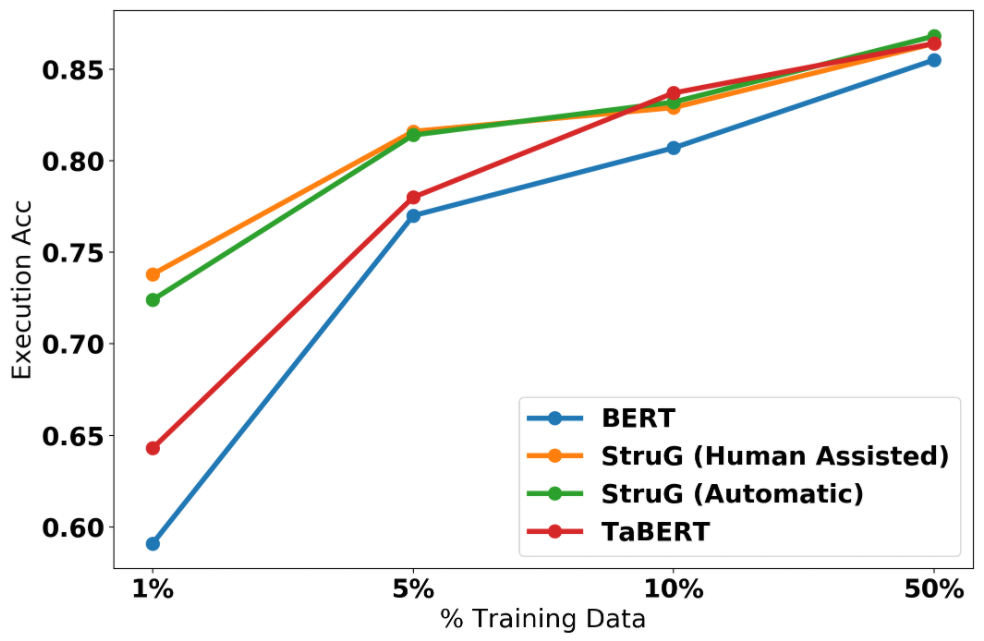}
\vspace{-5pt}
\caption{Execution Accuracy on the WikiSQL test set with different fractions of training data.}
\vspace{-10pt}
\label{fig:wikisql_x_change}
\end{figure}
\begin{figure}[t]
  \centering
  \includegraphics[width=0.75\linewidth]{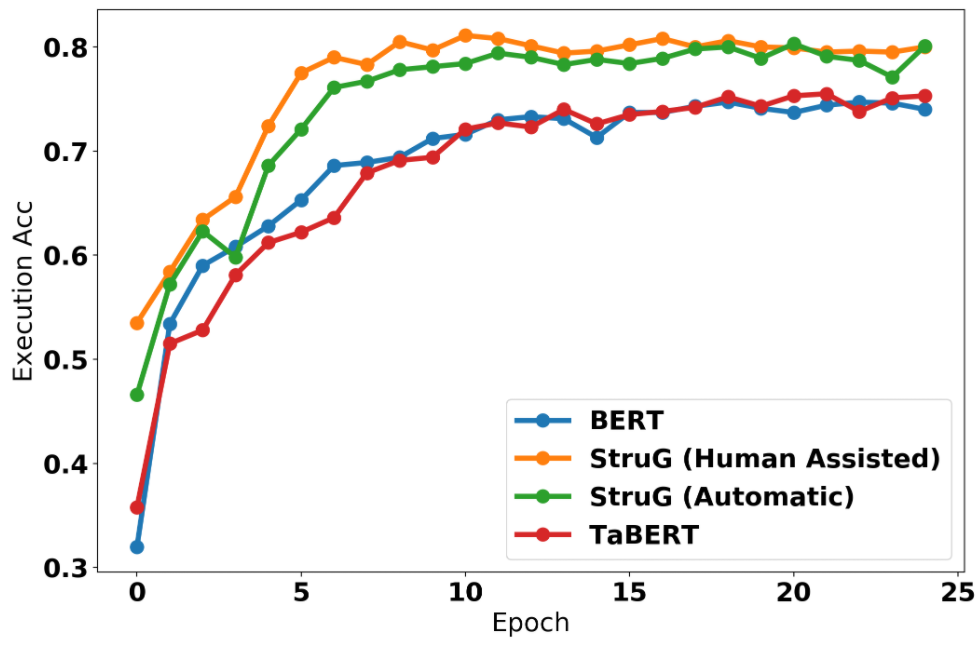}
 \vspace{-5pt}
\caption{Execution Accuracy on the WikiSQL dev set during training with 5\% of training data.}
\vspace{-10pt}
\label{fig:wikisql_x_train}
\end{figure}

\noindent\textbf{WikiSQL.}
Results on WikiSQL are summarized in \cref{tab:wikisql}. When using the full training corpus, we notice that using \ours \ achieves similar performance as BERT\textsubscript{LARGE}. This is probably because of the large size of training data and the simple SQL structure of WikiSQL. To better demonstrate that the knowledge learned in pretraining can be effectively transferred to text-to-SQL task and reduce the need for supervised training data, we also conduct experiments with randomly sampled training examples. From \cref{fig:wikisql_x_change} we can see that with only 1\% of training data (around 500 examples), models using \ours \ can achieve over 0.70 accuracy, outperforming both BERT\textsubscript{LARGE} and TaBERT\nop{Huan: do we need one-sentence note for why in some tables TaBERT is compared and why in others GRAPPA is compared?e.g., b/c we directly copy their numbers?} by a large margin. \ours\ brings consist improvement over BERT\textsubscript{LARGE} until we use half of the training data, where all models reach nearly the same performance as using the full training data. We also show the training progress using 5\% of training data in \cref{fig:wikisql_x_train}. We can see that \ours\ also helps speed up the training progress. For more break-down results on several subtasks, please see \cref{sec:more_results_wikisql}.

\noindent\textbf{Comparison of human assisted and automatic setting.}
In all benchmarks, we notice that \ours\ pretrained using the automatic setting actually performs similarly as the setting where cell annotations are used. This indicates the effectiveness of our heuristic for cell annotation and the potential to pretrain \ours \ with more unannotated parallel text-table data.

\begin{table}[]
    \centering
    \resizebox{0.85\linewidth}{!}{
    \begin{tabularx}{1.05\linewidth}{cl}
        \toprule
        \multirow{3}{*}{\rotatebox[origin=c]{90}{\small \hspace{-55pt}Spider-Realistic}}&\makecell[l]{What are the names of tournaments that have\\ more than 10 matches?}\\
        &\makecell[l]{\textbf{\small \textit{w.} \ours \ (Automatic)} \cmark\\ \uline{\texttt{\small SELECT \textcolor{blue}{tourney\_name} FROM matches}}\\ \uline{\texttt{\small GROUP BY \textcolor{blue}{tourney\_name}}}\\\uline{\texttt{\small HAVING Count(*) > 10}}}\\
        &\makecell[l]{\textbf{\small \textit{w.} BERT\textsubscript{LARGE}} \xmark\\\texttt{\small SELECT \textcolor{red}{first\_name} FROM players JOIN}\\ \texttt{\small matches GROUPBY \textcolor{red}{first\_name}}\\\texttt{\small HAVING Count(*) > 10}}\\
        \midrule
        \multirow{3}{*}{\rotatebox[origin=c]{90}{\small\hspace{-40pt} IMDB}}&\makecell[l]{List " James Bond " directors}\\
        &\makecell[l]{\textbf{\small \textit{w.} \ours \ (Automatic)} \cmark\\\uline{\texttt{\small SELECT \textcolor{blue}{name} FROM director}}\\\uline{\texttt{\small JOIN directed\_by JOIN MOVIE}}\\\uline{\texttt{\small WHERE \textcolor{blue}{movie.title} = "james bond"}}}\\
        &\makecell[l]{\textbf{\small \textit{w.} BERT\textsubscript{LARGE}} \xmark\\\texttt{\small SELECT \textcolor{red}{gender} FROM director}\\ \texttt{\small WHERE \textcolor{red}{director.name} = "james bond"}}\\
        \bottomrule
    \end{tabularx}}
    \caption{Case study.\nop{SQL queries with underline are correct queries generated by model \textit{w.} \ours \ (Automatic), the others are wrong predictions made by model \textit{w.} BERT\textsubscript{LARGE}.}}
    \label{tab:error}
    \vspace{-15pt}
\end{table}
\subsection{Case Study}
\label{sec:case}
\add{We compare the predictions made by RAT-SQL \textit{w.} BERT\textsubscript{LARGE} and \textit{w.} \ours \ (Automatic). Some examples are shown in \cref{tab:error}. In the first example from Spider-Realistic, we can see that the model \textit{w.} BERT\textsubscript{LARGE} fails to align \textit{tournaments} with the \textit{tourney\_name} column, because of string mismatch. In the second example from IMDB, although the model correctly recognizes \textit{James Bond} as value reference, it fails to ground it to the correct column which is \textit{movie\_title}. This supports our hypothesis that using \ours\ helps to improve the structure grounding ability of the model.}

\vspace{-5pt}
\section{Conclusion}
\vspace{-5pt}
In this paper, we propose a novel and effective structure-grounded pretraining technique for text-to-SQL. Our approach to pretraining leverages a set of novel prediction tasks using a parallel text-table corpus to help solve the text-table alignment challenge in text-to-SQL. We design two settings to obtain pretraining labels without requiring complex SQL query annotation: using human labeled cell association, or leveraging the table contents. In both settings, \ours \ significantly outperforms BERT\textsubscript{LARGE} in all the evaluation sets. Meanwhile, although \ours \ is surprisingly effective (using only 120k text-table pairs for pretraining) and performs on par with models like TaBERT (using 26m tables and their English contexts) and GRAPPA (using 475k synthetic examples and 391.5k examples from existing text-table datasets) on Spider, \nop{Huan: although \ours \ is surprisingly effective (using only 120k text-table pairs for pretraining) and performs on par with models like TaBERT and GRAPPA (both using XX text-table pairs),} we believe it is complementary with these existing text-table pretraining methods. In the future, we plan to further increase the size of the pretraining corpus, and explore how to incorporate MLM and synthetic data.
\section*{Ethical Considerations}
\noindent\textbf{Dataset.}
In this work, we re-purpose an existing table-to-text generation dataset ToTTo \cite{parikh2020totto} for our pretraining. We obtain labels for our three pretraining tasks via weak supervision, which uses only the raw sentence-table pairs, or the cell annotations and revised descriptions that are already included in ToTTo dataset. As a result, no extra human effort is required for collecting our pretraining corpus. We also curate a more realistic evaluation dataset for text-to-SQL based on Spider dev set. In particular, we first select a complex subset from the Spider dev set and manually revise the NL questions to remove the explicit mention of column names. The detailed description of the process can be found in \cref{sec:new_eva}. The first author manually revised all the questions himself, which results in 508 examples in total.

\noindent\textbf{Application.}
We focus on the task of text-to-SQL, which is a fundamental task for building natural language interfaces for databases. Such interface can enable non-expert users to effortlessly query databases. In particular, here we focus on improving the structure grounding ability of text-to-SQL models, which is critical in real-world use cases. We evaluate our model with the widely used Spider benchmark and several more realistic datasets. Experimental results show that our method brings significant improvement over existing baselines, especially on more realistic settings.

\noindent\textbf{Computing cost.} 
We use 4 V100 GPUs for pretraining, and 1 V100 GPU for finetuning the model for text-to-SQL on Spider and WikiSQL. One advantage of our method is its efficiency. In our experiments, we pretrain the model for only 5 epochs, which can finish within 4 hours. For comparison, the largest TaBERT model \cite{yin2020tabert} takes 6 days to train for 10 epochs on 128 Tesla V100 GPUs using mixed precision training.
\section*{Acknowledgements}
We thank Bo Pang, Tao Yu for their help with the official Spider evaluation. We also thank anonymous reviewers for their constructive feedback.

\bibliography{ref}
\bibliographystyle{acl_natbib}
\clearpage
\appendix
\section{Implementation Details}
\begin{table}[t]
    \centering
    \resizebox{\linewidth}{!}{
    \begin{tabular}{lccccc}
    \toprule
        Models&Easy&Medium&Hard&Extra Hard&All\\
        \midrule
        \# Examples &248&446&174 &166&1034\\
        \midrule
        RAT-SQL \textit{w/o} value linking &&& &&\\
        \hspace{10pt}\textit{w.} BERT\textsubscript{LARGE} &82.9&72.7&65.7&46.6&69.8\\
        \hspace{10pt}\textit{w.} \ours \ (Human Assisted)  &84.9&76.2&67.0&55.0&73.3\\
        \hspace{10pt}\textit{w.} \ours \ (Automatic)  &87.8&75.6&69.5&55.0&74.2\\
        \midrule
         RAT-SQL &&&&&\\
        \hspace{10pt}\textit{w.} BERT\textsubscript{LARGE}  &84.1&74.9&67.4&52.8&72.3\\
        \hspace{10pt}\textit{w.} \ours \ (Human Assisted)  &87.1&77.7&70.9&57.0&75.5\\
        \hspace{10pt}\textit{w.} \ours \ (Automatic)  &88.7&77.4&69.2&53.6&74.9\\
        \bottomrule
        
    \end{tabular}
    }
    \caption{Execution accuracy on Spider dev set with different hardness levels.}
    \label{tab:spider_difficulty}
\end{table}
\begin{table}[t]
    \centering
    \resizebox{\linewidth}{!}{
    \begin{tabular}{lcccc}
    \toprule
        Models&\texttt{SELECT}&\texttt{WHERE}&\texttt{GROUP BY}&\texttt{ORDER BY}\\
        \midrule
        RAT-SQL \textit{w/o} value linking &&& &\\
        \hspace{10pt}\textit{w.} BERT\textsubscript{LARGE} &89.2&71.7&78.7 &81.5\\
        \hspace{10pt}\textit{w.} \ours \ (Human Assisted)  &91.2&74.8&79.0 &84.0\\
        \hspace{10pt}\textit{w.} \ours \ (Automatic)  &90.9&75.6&77.5 &84.0\\
        \midrule
         RAT-SQL &&& &\\
        \hspace{10pt}\textit{w.} BERT\textsubscript{LARGE}  &89.4&79.2&78.5 &81.3\\
        \hspace{10pt}\textit{w.} \ours \ (Human Assisted)  &91.3&80.8&80.6 &85.7\\
        \hspace{10pt}\textit{w.} \ours \ (Automatic)  &91.2&80.1&78.6 &84.5\\
        \bottomrule
        
    \end{tabular}
    }
    \caption{ F1 scores of Component Matching on Spider dev set.}
    \vspace{-10pt}
    \label{tab:spider_sub_0}
\end{table}
\begin{table}[t]
    \centering
    \resizebox{\linewidth}{!}{
    \begin{tabular}{lcccc}
    \toprule
        Models&\texttt{SELECT}&\texttt{WHERE}&\texttt{GROUP BY}&\texttt{ORDER BY}\\
        \midrule
        RAT-SQL \textit{w/o} value linking &&& &\\
        \hspace{10pt}\textit{w.} BERT\textsubscript{LARGE} &86.2&55.6&65.9&64.3\\
        \hspace{10pt}\textit{w.} \ours \ (Human Assisted)  &88.9&61.9&70.4&64.1\\
        \hspace{10pt}\textit{w.} \ours \ (Automatic)  &90.1&64.5&73.0&67.4\\
        \midrule
         RAT-SQL &&& &\\
        \hspace{10pt}\textit{w.} BERT\textsubscript{LARGE}  &86.9&74.2&59.6&61.9\\
        \hspace{10pt}\textit{w.} \ours \ (Human Assisted)  &89.0&76.8&69.9&63.5\\
        \hspace{10pt}\textit{w.} \ours \ (Automatic)  &89.2&76.4&64.7&64.9\\
        \bottomrule
    \end{tabular}
    }
    \caption{ F1 scores of Component Matching on \newdata set.}
    \vspace{-10pt}
    \label{tab:spider_sub_1}
\end{table}
% \subsection{More Details on Preparing the Pretraining Data}
% \label{sec:pretrain_data_detail}
\subsection{Filtering on the \citet{suhr2020exploring} Datasets}
\label{sec:data_filter}
We use the filtering scripts\footnote{\url{https://github.com/google-research/language/tree/master/language/xsp}} released by the authors of \citet{suhr2020exploring}. More specifically, they remove examples that fall into the following categories: (1) a numeric or text value in the query is not copiable from the utterance (except for the numbers 0 and 1, which are often not copied from the input), (2) the result of the query is a empty table, or a query for count returns \texttt{[1]}, (3) the query requires selecting more than one final column.

\begin{figure*}
\centering
\begin{subfigure}{.4\linewidth}
  \centering
  \includegraphics[width=\linewidth]{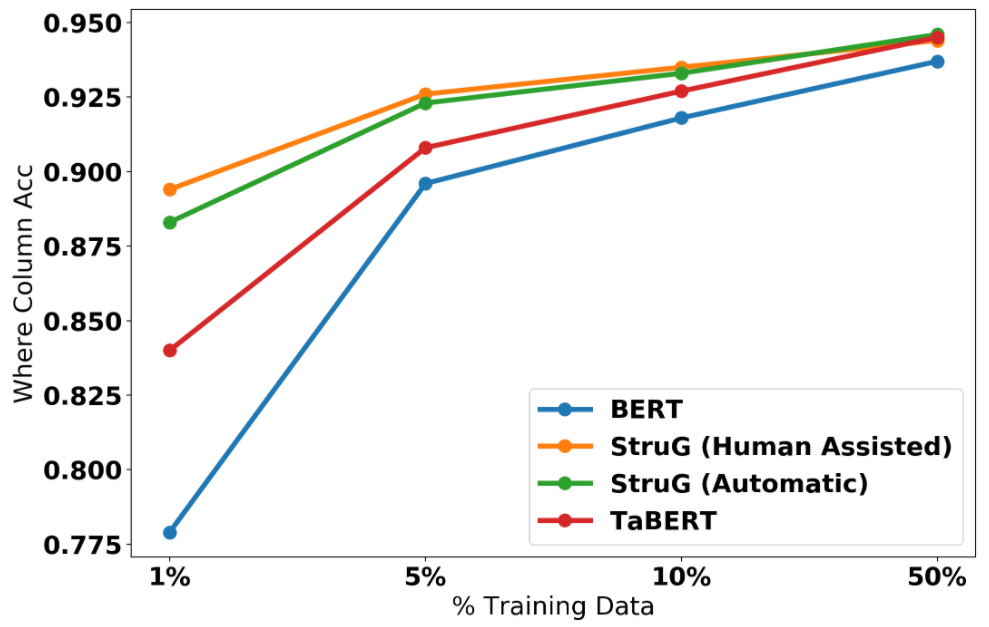}
  \caption{Where Column Accuracy}
  \label{fig:wc_change}
\end{subfigure}
~
\begin{subfigure}{.4\linewidth}
  \centering
  \includegraphics[width=\linewidth]{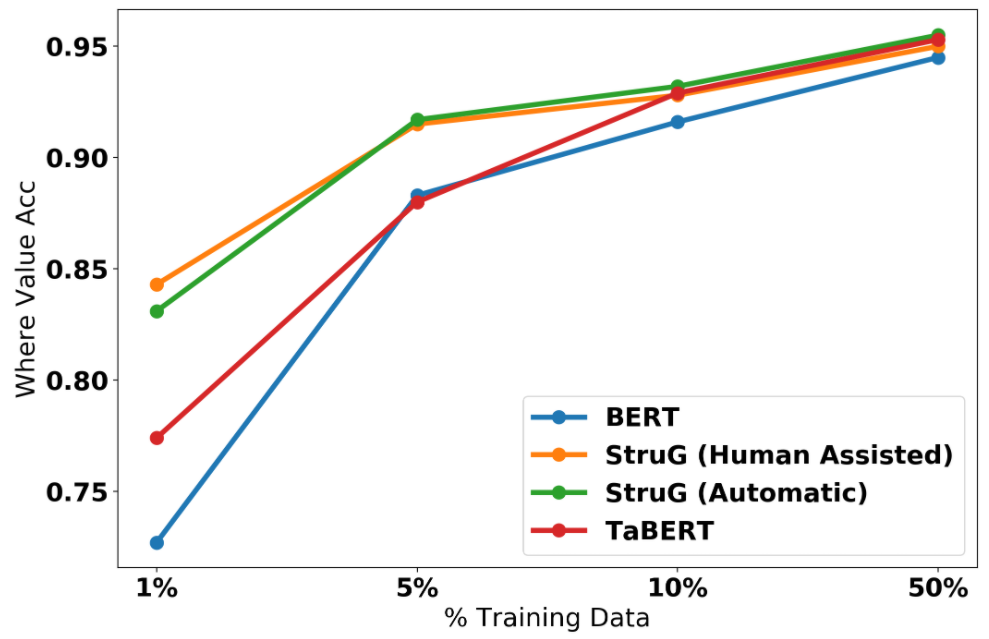}
  \caption{Where Value Accuracy}
  \label{fig:wv_change}
\end{subfigure}
\vspace{-5pt}
\caption{Model performance on the test set with different fractions of training data.}
\label{fig:wikisql_change}
\end{figure*}

\begin{figure*}
\centering
\begin{subfigure}{.4\linewidth}
  \centering
  \includegraphics[width=\linewidth]{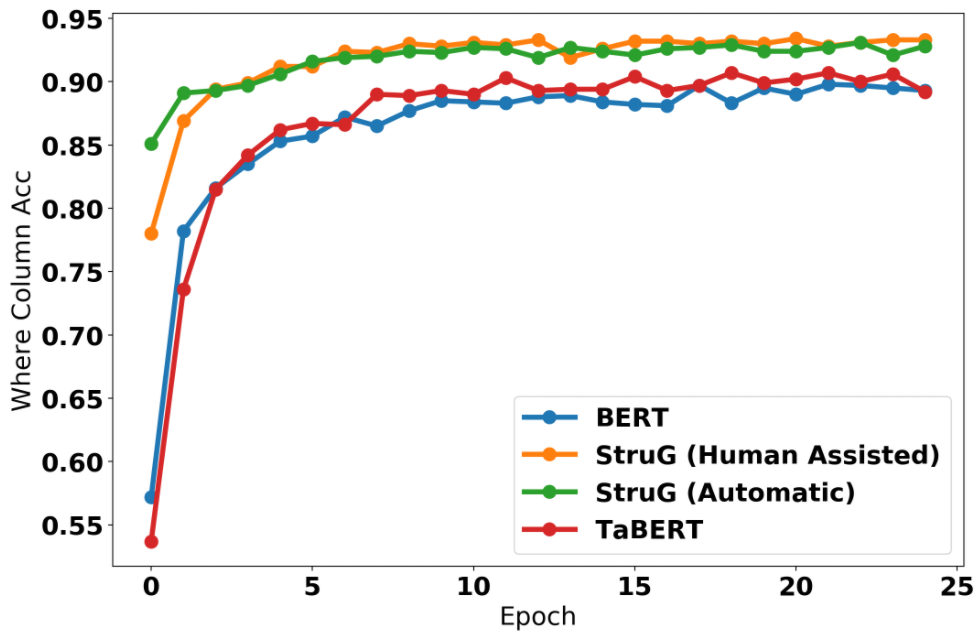}
  \caption{Where Column Accuracy}
  \label{fig:wc_train}
\end{subfigure}
~
\begin{subfigure}{.4\linewidth}
  \centering
  \includegraphics[width=\linewidth]{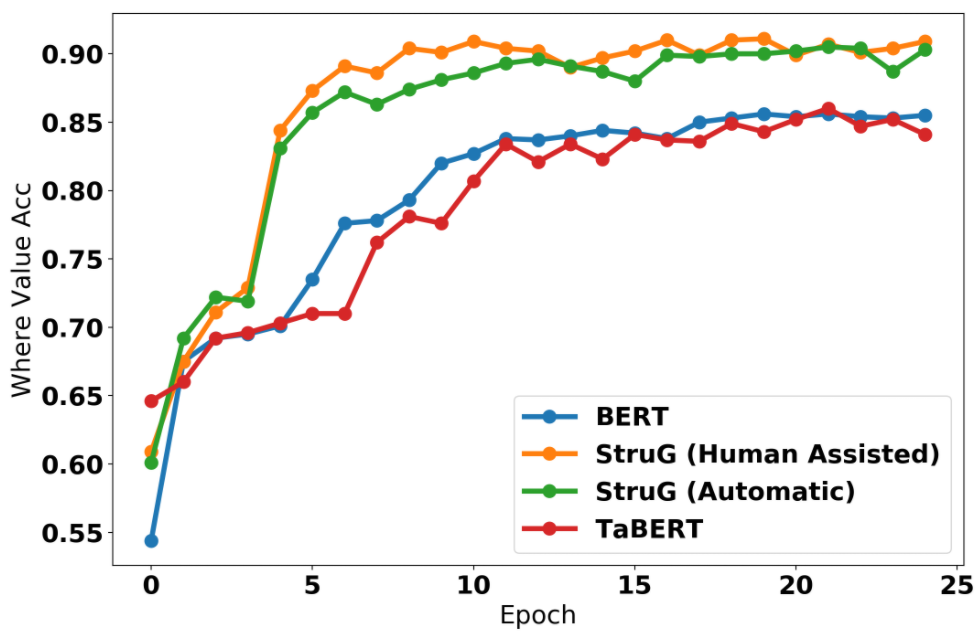}
  \caption{Where Value Accuracy}
  \label{fig:wv_train}
\end{subfigure}
\vspace{-5pt}
\caption{Model performance on the dev set during training with 5\% of training data.}
\vspace{-10pt}
\label{fig:wikisql_train}
\end{figure*}
\section{More Resutls}
\subsection{Detailed Results on Spider and \newdata}
\label{sec:more_results_spider}
We show more detailed results on the Spider dev set and Spider-Realistic in \cref{tab:spider_difficulty}, \cref{tab:spider_sub_0} and \cref{tab:spider_sub_1}. From \cref{tab:spider_difficulty} we can see that \ours\ brings significant improvements in all difficulty levels, and is not biased towards certain subset. Since \ours\ mostly improves the structure grounding ability of the model, from \cref{tab:spider_sub_0} and \cref{tab:spider_sub_1}, we can see that \ours\ mainly increase the accuracy for \texttt{WHERE} and \texttt{ORDER BY} clauses, especially when database content is not available to the model. On the \newdata set, as the model cannot rely on simple string matching for structure grounding, we notice greater improvement using \ours, especially for \texttt{GROUP BY} clauses.
\subsection{Detailed Results on WikiSQL}
\label{sec:more_results_wikisql}
\begin{table}[]
    \centering
    \resizebox{\linewidth}{!}{
    \begin{tabular}{lcccc}
    \toprule
        Models & ACC\textsubscript{S-COL}& ACC\textsubscript{S-AGG}& ACC\textsubscript{W-COL}& ACC\textsubscript{W-VAL} \\
        \midrule
        % SQLova \cite{hwang2019sqlova} &&&& \\
        % \hspace{10pt}\textit{w.} BERT\textsubscript{LARGE} &97.3 &91.0 &94.8&95.8 \\
        % \hspace{10pt}\textit{w.} TaBERT &97.0 &90.9 &95.3&96.2 \\
        % \hspace{10pt}\textit{w.} \ours \ (Human Assisted) &97.1 &90.9 &95.1&95.8 \\
        % \hspace{10pt}\textit{w.} \ours \ (Automatic) &96.9&90.5&95.1&95.9 \\
        % \midrule
        SQLova (5\%) & & && \\
        \hspace{10pt}\textit{w.} BERT\textsubscript{LARGE} &95.2 &88.4 &89.6&88.3 \\
        \hspace{10pt}\textit{w.} TaBERT &95.4 &88.4 &90.8&88.0 \\
        \hspace{10pt}\textit{w.} \ours \ (Human Assisted) &95.5 &88.9 &92.6&91.5 \\
        \hspace{10pt}\textit{w.} \ours \ (Automatic) &95.8 &88.9 &92.3&91.7 \\
         \bottomrule
    \end{tabular}}
    \caption{Subtask performance on WikiSQL. S-COL, S-AGG, W-COL and W-VAL stands for tasks of predicting SELECT column, aggregation operator, WHERE columns and WHERE values, respectively.}
    \label{tab:wikisql_detailed}
\end{table}
We show subtask performance for WikiSQL in \cref{tab:wikisql_detailed}, \cref{fig:wikisql_train} and \cref{fig:wikisql_x_change}. Again, we can see that \ours\ mainly improves \texttt{WHERE} column and \texttt{WHERE} value accuracy. From \cref{fig:wikisql_change} we can see that with only 1\% of training data, model with \ours\ already has over 0.87 \texttt{WHERE} column accuracy and nearly 0.85 \texttt{WHERE} value accuracy.
\end{document}